\documentclass[sigconf]{acmart}

\usepackage{algorithm}
\usepackage{algorithmic}
\usepackage{graphicx}
\usepackage{amsmath}
\usepackage{breqn}
\usepackage{listings}
\usepackage{pifont}

\AtBeginDocument{%
  }

\setcopyright{rightsretained}
\copyrightyear{2024}
\acmYear{2024}
\acmDOI{N.A}

\acmConference{Preprint}{Working Paper}{Nov 2024}
\begin{document}

% \title{CAML - Collaborative LLM-Agents with Multi-task Contrastive Learning for Multi-Turn Intent Classification}
% \title{Inferring Chain-of-Intent by Collaborative LLM-Agents for Multi-Turn Intent Generation}
% \title{Inferring Chain-of-Intent by LLM-Agentic Workflow for Multi-Turn Intent Recognition}
\title{Balancing Accuracy and Efficiency in Multi-Turn Intent Classification for LLM-Powered Dialog Systems in Production}

\author{
Junhua Liu\textsuperscript{1,3,$^*$}, 
Yong Keat Tan\textsuperscript{2,$^*$},
Bin Fu\textsuperscript{2,$^\dagger$},
Kwan Hui Lim\textsuperscript{3}
}
\affiliation{
 \institution{\textsuperscript{1}Forth AI}
 \institution{\textsuperscript{2}Shopee}
 \institution{\textsuperscript{3}Singapore University of Technology and Design}
 \country{Singapore}
}

% \affiliation{%

\renewcommand{\shortauthors}{Liu et al.}

\begin{abstract}
Accurate multi-turn intent classification is essential for advancing conversational AI systems. However, challenges such as the scarcity of comprehensive datasets and the complexity of contextual dependencies across dialogue turns hinder progress. This paper presents two novel approaches leveraging Large Language Models (LLMs) to enhance scalability and reduce latency in production dialogue systems. First, we introduce Symbol Tuning, which simplifies intent labels to reduce task complexity and improve performance in multi-turn dialogues. Second, we propose C-LARA (Consistency-aware, Linguistics Adaptive Retrieval Augmentation), a framework that employs LLMs for data augmentation and pseudo-labeling to generate synthetic multi-turn dialogues. These enriched datasets are used to fine-tune a small, efficient model suitable for deployment. Experiments conducted on multilingual dialogue datasets demonstrate significant improvements in classification accuracy and resource efficiency. Our methods enhance multi-turn intent classification accuracy by 5.09\%, reduce annotation costs by 40\%, and enable scalable deployment in low-resource multilingual industrial systems, highlighting their practicality and impact.
\end{abstract}





\maketitle
\def\thefootnote{*}\footnotetext{Equal Contributions.}
\def\thefootnote{$\dagger$}\footnotetext{Corresponding Author: \texttt{bin.fu@shopee.com}}

\section{Introduction}

Dialogue systems are critical for automating interactions between customers and agents, streamlining communication and enhancing user experience. They play a pivotal role in international e-commerce platforms, addressing the increasing demand for instantaneous and efficient customer service. Intent classification, a fundamental aspect of natural language understanding in dialogue systems, involves identifying users’ goals from their inputs, thereby minimizing waiting times and operational costs~\cite{weld2021survey}. User interactions frequently evolve into multi-turn dialogues when detailed information is required, complicating the development of multi-turn intent classification (MTIC) models, despite their similarity to standard text classification tasks. Additionally, real-world multilingual systems require scalable solutions that uphold inclusivity and ethical standards, particularly in low-resource settings. This complexity arises from the need to consider contextual factors like historical utterances and prior intents. Without a proper understanding of session context, the system risks misinterpreting user intentions, which may result in incorrect applications or irrelevant responses \cite{6853573}. Consequently, MTIC within dialogue system presents significant challenges.

\begin{figure}[t]
    \centering
    \includegraphics[width=.45\textwidth]{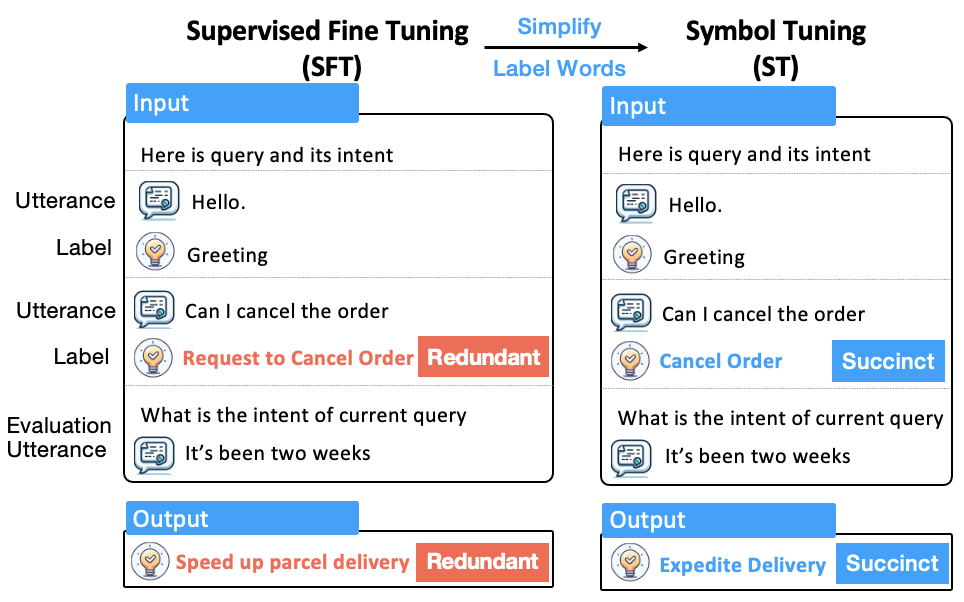}
    \vspace{-.3cm}
    \caption{Comparison of instruction tuning and symbol tuning. Simplifying verbose intent labels (e.g., “Request to Cancel Order” → “Cancel Order”) reduces redundancy, enhancing LLM classification performance by 5.09\%, addressing key challenges in production intent classification.}
    \label{fig:symboltuning}
    \vspace{-.5cm}
\end{figure}

The first challenge is that the length of intents in industrial dialogue systems is longer compared to general text classification tasks. Figure ~\ref{fig:symboltuning} shows that the real intents comprise several words in our knowledge base because operators(Ops) typically assign intents a clear and descriptive name to facilitate knowledge management, which makes them redundant. The recent advancements in large language model(LLMs) present new research opportunities to simplify and optimize the text classification process \cite{wang2024smart}. Research indicates that LLMs perform excellently in sentiment analysis\cite{pvribavn2024comparative}, which only adopts shorter labels such as positive, negative. However, LLMs still fail to address context dependency in multi-turn conversations and struggle with long intent labels common in industrial systems.

\begin{figure}[t]
    \centering
    \includegraphics[width=.45\textwidth]{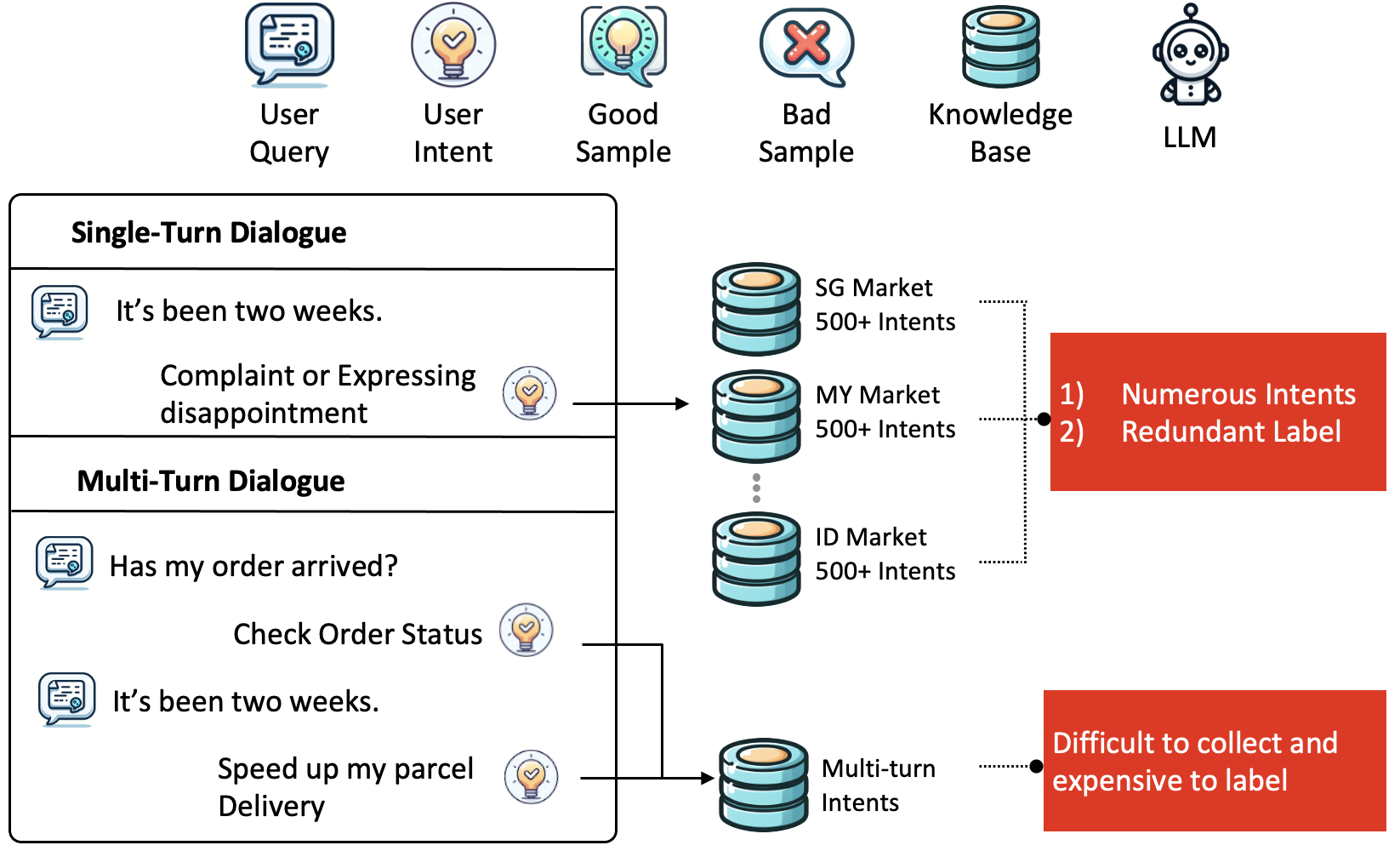}
    \vspace{-.2cm}
    \caption{Annotation pipeline of multi-turn intent classification datasets. Two major challenges in production systems are illustrated: (1) managing numerous (500+) intents across markets with redundant labels, and (2) the high cost of collecting multi-turn training data.}
    \label{fig:challenges}
    \vspace{-.5cm}
\end{figure}

The second challenge lies in the difficulty of collecting multi-turn datasets. While several studies~\cite{Qu_2019, wu2021contextaware} on MTIC exist, they often assume access to comprehensive multi-turn training data, which is rarely available in real-world applications.

Figure ~\ref{fig:challenges} shows the annotation pipeline for MTIC tasks. Even if we ignore the redundant info within intents, unlike dialogue act classification \cite{qin2020dcr} with only less than 10 classes within dialogue state tracking (DST), there are hundreds of intents operated by local Ops in knowledge base of dialgoue system to cover user's various and specific intents in each market, which increase the complexity of multi-turn classification and its annotation. Annotators often struggle with the numerous of intents, leading to increased mistakes and longer decision-making times. As a result, the annotation process becomes costly and time-intensive, making it impractical to manually annotate large-scale multi-turn datasets. However, insufficient training data can significantly hinder model performance even with LLM. These challenges underscore the necessity for more efficient methods to address data scarcity and classification complexity.

To tackle two challenges, we first study the feasibility of using LLM for supervised fine-tuning (SFT) to perform MTIC using a generative method. Various intents increase the complexity of this task since the more tokens a large language model(LLM) generates, the lower the task performance \cite{rust2021good}. To conquer this, we compress the redundant info within intent to succinct intent via GPT4, then adopt those intents in SFT named as symbol tuning, which help to reduce the difficulty of multi-turn classification tasks by the LLM generative method. 

Secondly, to overcome the shortage of multi-turn data, we propose a novel pseudo-labeling and data generation framework called Consistency-aware Linguistics Adaptive Retrieval Augmentation (C-LARA). Extending beyond existing synthetic data generation~\cite{junhua2024lara}, C-LARA serves as an effective pseudo-labeling tool for generating multi-turn data from user's unlabeled utterances with self-consistency. C-LARA arranges the retrieval result in different orders to assemble adaptive prompts, which cover the diverse reasoning path and filter out noise in in-context learning to improve the quality of labeling data. Subsequently,  we use the training data to train a smaller model for online inference. C-LARA is a novel framework tailored for multi-turn intent classification. It addresses limitations in prior approaches by leveraging adaptive retrieval and self-consistency mechanisms to enhance the accuracy of pseudo-labeling for multi-turn dialogues. Unlike previous methods, it directly optimizes for zero-shot multi-turn data classification and scalable deployment.

In summary, the contributions of this paper are as follows:

\begin{enumerate}
  \item We introduce symbol-tuning, leveraging compressed intents to enhance LLM performance for MTIC, demonstrating a 5.09\% improvement in supervised fine-tuning (SFT) results.
  \item We develop C-LARA, a novel framework for generating high-quality multi-turn data, effectively augmenting MTIC results. 
  \item We fine-tune smaller models using data generated by C-LARA, enabling scalable and accurate deployment of MTIC systems in low-resource industrial settings.
\end{enumerate}
\section{Problem Formulation}
% problem formulation..

%  - copy from lara
%    - introduce multi-turn conversation
%    - finetuning + ICL

% % Solution
% % 1. finetuning LLM (intent shortening, cross-lingual label, different base model - in experiment setting)
% % 2. LARA (simple introduction and cite)
% % 3. pseudo-labeling (plus self-consistency) + traditional model

% % Experiment
% % 1. self-consistency (traditional model) here

% % Ablation
% % 1. self-consistency (precision vs accuracy on ICL)
% % 2. the shorter the answer, the better, but should keep the semantic meaning https://docs.google.com/spreadsheets/d/1XEqKLYYvwBkeVNotM3rKtdRRyo9ZZ4epjl0nzONj7ds/edit?gid=0#gid=0&range=B5:C8
% % 3. prefix lm (https://arxiv.org/pdf/2410.17532) https://docs.google.com/spreadsheets/d/1XEqKLYYvwBkeVNotM3rKtdRRyo9ZZ4epjl0nzONj7ds/edit?gid=0#gid=0&range=C17:G18

\subsection{Multi-Turn Intent Classification}
Multi-Turn Intent Classification (MTIC) involves identifying the intent \( I \) of the final query \( q_n \) from a predefined set \( \mathcal{I} \), based on a sequence of user queries \( \mathcal{Q} = \{q_i\}_{i=1}^n \) in a chatbot session. This task relies on the conversational context \( \mathcal{C} = \{q_i\}_{i=1}^{n-1} \), which includes prior queries. Context-dependency adds complexity, requiring models to interpret nuanced conversational dynamics and evolving user intentions. Each intent \( I \) has a local-language title \( y \) and a hierarchical English category \( z \) (e.g., Indonesia: \( y = \text{'Cara membatalkan pesanan'} \), \( z = \text{'Logistics > Order > Cancellation'} \)).

\subsection{Supervised Fine-tuning}
Supervised Fine-tuning (SFT) adapts pre-trained large language models (LLMs) for specific tasks using labeled datasets. This process achieves high benchmark accuracy through task-specific supervision.

\subsubsection*{Problem Definition}
Given a dataset \( \mathcal{D} = \{(x_i, y_i)\}_{i=1}^N \), where \( x_i \) is an input query and \( y_i \) is the corresponding label, the objective is to optimize model parameters \( \theta \) to maximize the conditional likelihood \( p(y_i | x_i; \theta) \):
\[
\mathcal{L}_{\text{SFT}}(\theta) = -\frac{1}{N} \sum_{i=1}^N \log p(y_i | x_i; \theta).
\]

\subsubsection*{Conditional Probability Modeling}
For structured outputs, \( y_i \) is a sequence of tokens \( \{t_1, t_2, \ldots, t_m\} \), with probability factorized autoregressively:
\[
p(y|x; \theta) = \prod_{j=1}^m p(t_j | t_{<j}, x; \theta).
\]
The training objective becomes:
\[
\mathcal{L}_{\text{SFT}}(\theta) = -\frac{1}{N} \sum_{i=1}^N \sum_{j=1}^m \log p(t_j | t_{<j}, x_i; \theta).
\]

\subsection{Symbol Tuning}
Unlike methods replacing task labels with unrelated symbols~\cite{wei2023symboltuningimprovesincontext}, our Symbol Tuning approach focuses on intent classification. Verbose labels in industrial systems disperse semantic information, hindering model performance. To address this, we compress labels into concise phrases using GPT-4. For example, "Request to Cancel Order" becomes "Cancel Order," serving as compact semantic anchors that enhance shallow and deep layer representations.

\subsubsection*{Mathematical Formulation}
Let the original intent label be \( L = \{t_1, t_2, \ldots, t_m\} \). The compressed label \( L' \), with \( n \ll m \), is generated by optimizing:
\[
L' = \operatorname{argmin}_{L'} \; \mathcal{C}(L') + \mathcal{E}(L', L),
\]
where:
- \( \mathcal{C}(L') \): Compactness of \( L' \) (e.g., token count).
- \( \mathcal{E}(L', L) \): Semantic divergence, computed as:
\[
\mathcal{E}(L', L) = 1 - \operatorname{cosine\_sim}(\phi(L'), \phi(L)),
\]
with \( \phi(\cdot) \) as an embedding function.

\subsubsection*{Objective Function}
Given \( \mathcal{D} = \{(x_i, L_i)\}_{i=1}^N \), where \( L_i \) is the original label, the supervised fine-tuning loss becomes:
\[
\mathcal{L}_{\text{ST}}(\theta) = -\mathbb{E}_{(x, L') \sim \mathcal{D}} \sum_{j=1}^n \log p(t_j | t_{<j}, x; \theta),
\]
where \( t_{<j} \) denotes preceding tokens in \( L' \).

\subsubsection*{Performance Implications}
Replacing verbose labels \( L \) with compact \( L' \) reduces token processing and improves classification accuracy, streamlining intent recognition tasks.

% \subsection{In-context Learning}
% The scaling of language models has enabled a variety of new applications and paradigms in machine learning, including the ability to perform challenging reasoning tasks via a few examples given in-context \cite{wei2023symboltuningimprovesincontext}. The In-context Learning (ICL) process does not require any weight update of a pre-trained LLM. However, it often requires careful prompt engineering to ensure the quality of outputs. 

\section{Solutions}
\subsection{Symbol Tuning on LLM}
To address intent classification tasks, we utilize \textbf{generative models} rather than conventional discriminative or regressive approaches. Our Symbol Tuning (ST) method involves supervised fine-tuning (SFT) of an LLM with compressed intent labels. Given a complete chat session \( \mathcal{S} = \{q_1, I_1, ..., q_{n-1}, I_{n-1}, q_n\} \), the model is trained to generate the representative question \( r_n \) corresponding to the correct intent \( I_n \) of the final query \( q_n \). Queries and intents are structured in a natural question-answering flow, as illustrated below:
\begin{quote}
    \textbf{SYSTEM}: "A chat between a curious user and an artificial intelligence assistant. The assistant provides helpful, detailed, and polite responses to the user's questions."\\
    \textbf{USER}: "\{q_1\}"\\
    \textbf{ASSISTANT}: "The intent title is \{r_1\}."\\
    ...\\
    \textbf{USER}: "\{q_n\}"\\
    \textbf{ASSISTANT}: "The intent title is \{r_n\}."
\end{quote}
The generated \( r_n \) is compared with intents in \( \mathcal{I} \) using cosine similarity in the embedding space to ensure semantic alignment between the model's output and predefined intent titles.

\subsubsection*{Compressed Generation} 
Intent representative queries $r$ often consist of approximately 12 tokens, making them inefficient as generation targets. To address this, we employ an LLM to compress $r$ into concise phrases, typically two words, while preserving their semantic essence. This process ensures that each compressed intent label $r_c$ is unique. If duplicates occur, the model iteratively increases the word count until uniqueness is achieved. This compression reduces the average length of $r_c$ to four tokens, optimizing it for generation tasks and improving classification accuracy. This approach enhances classification accuracy by reducing semantic dispersion in labels, ensuring more focused information propagation through LLM layers.

\subsubsection*{Cross-Lingual Labels} 
In non-English markets, intent labels \( r \) are compressed into English while retaining the original language for input queries \( \mathcal{Q} \). Leveraging English, the predominant language in LLM pretraining corpora, simplifies label generation and enhances model performance in multilingual settings. This cross-lingual strategy reduces complexity and improves alignment with pretraining distributions. This strategy leverages the strengths of pre-trained LLMs while accommodating multilingual data, offering a scalable solution for cross-lingual intent classification.

\begin{figure*}[t]
    \centering
    \includegraphics[width=.98\textwidth]{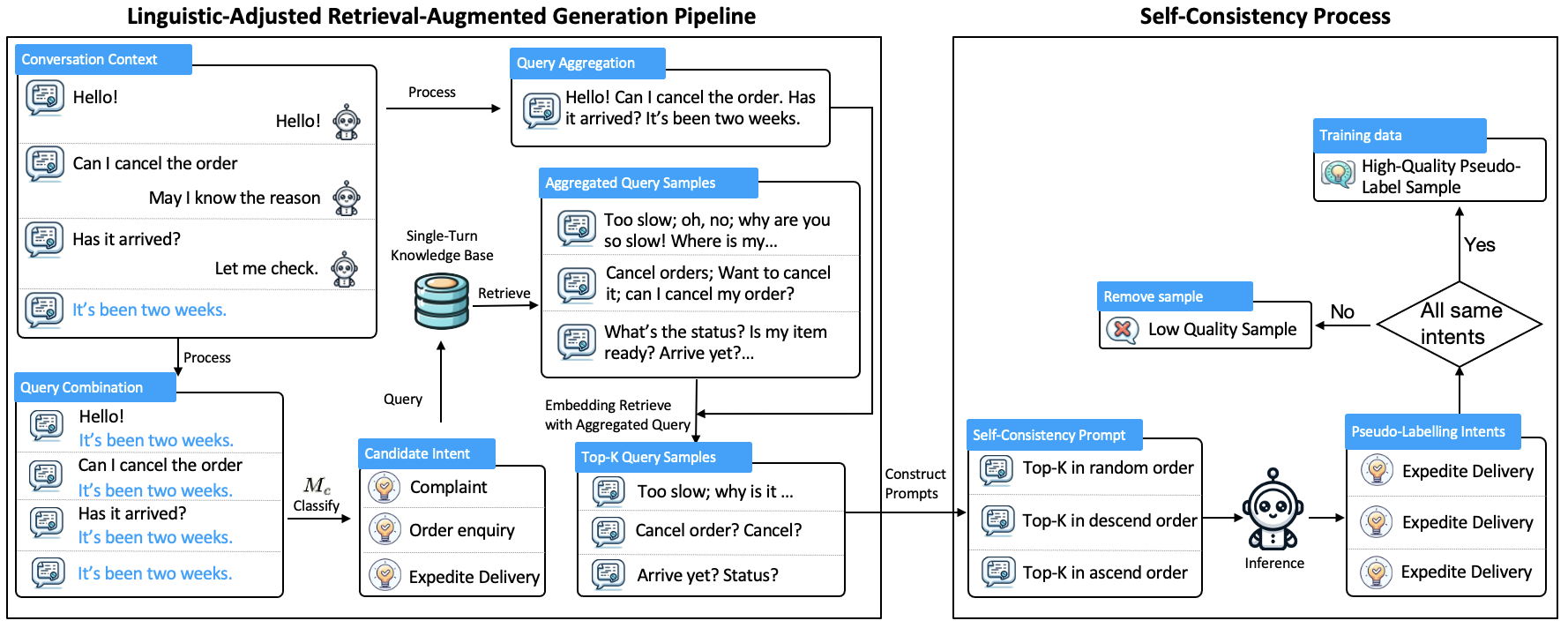}
    \vspace{-.3cm}
    \caption{Illustration of C-LARA: merging LARA with Self-Consistency effectively combines query aggregation, knowledge base retrieval, and self-consistency mechanism to generate high-quality pseudo-labels for multi-turn dialogues. The self-consistency process improves labeling accuracy by validating intent predictions across different prompt orderings.}
    \label{fig:clara-pipeline}
    \vspace{-.3cm}
\end{figure*}

\subsection{ Consistency-aware Linguistics Adaptive Retrieval Augmentation }
To enhance in-context learning, we propose the Consistency-aware, Linguistics Adaptive Retrieval Augmentation (C-LARA) framework. Building upon the LARA model~\cite{junhua2024lara}, C-LARA incorporates a fine-tuned single-turn model  $\mathcal{M}_c$ within a retrieval-augmented pipeline. This framework enables zero-shot Multi-Turn Intent Classification (MTIC) using only single-turn demonstrations. Unlike LARA, which is computationally intensive in real-time, C-LARA operates offline as a pseudo-labeling tool, generating high-quality multi-turn data for training lightweight classification models.

Specifically, the LARA pipeline can be complex and resource-intensive to implement for real-time systems. Hence, we use this method offline as a multi-turn data pseudo-labeling tool to train a smaller classification model. The training method mirrors that of the single-turn classifier $\mathcal{M}_c$ following the paper, adding pseudo-labeled multi-turn data to the original data comprising only single-turn samples. 

Since this is not a real-time task, the pipeline response time is not a critical consideration, hence self-consistency checking was performed on the LLM outputs to ensure the quality of pseudo-labels. For this check, as shown in Figure ~\ref{fig:clara-pipeline}, the in-context learning phase is run three times per sample, with the in-context demonstrations sorted in three orders according to their similarity scores to the session queries: ascending, descending, and random. This approach to self-consistency checking method can also be implemented when using a black-box LLM. Online chat logs are sampled for pseudo-labeling, and only those having consistent labels for all 3 runs will be kept for training. 

\subsubsection{Hierarchical Text Classification(HTC)}
$\mathcal{M}_c$ is an ensemble of label-attention encoder and a hierarchical-aware tree-based encoder with 3-layered global and local intent classifiers.

The label-attention encoder has one classifier head for each intent layer. Each classifier head has one hidden linear layer to obtain the layer intermediate output $L_l$, which encodes the layer information. This layer information will be utilised in the input of the next layer classifier head. 

\[
    L_l = 
    \begin{cases}
        HW^1_l + b^1_l, & \text{if } l = 1, \\
        (H \oplus L_{l-1})W^1_l + b^1_l, & \text{if } l > 1,
    \end{cases}
\]
where \( W^1_l \in \mathbb{R}^{d \times d} \text{ for } l = 1 \text{ and } W^1_l \in \mathbb{R}^{2d \times d} \text{ for } l > 1 \). $b^1_l \in \mathbb{R}^d$, $l$ is the layer number, $\oplus$ denotes tensor concatenation. Finally, we obtain the local logits $H_{local}^l$ for each layer classes by using another linear layer
$$ H_{local}^l = L_l \cdot W^2_l + b^2_l, W^2_l \in \mathbb{R}^{d \times |\mathcal{I}_l|}, b^2_l \in \mathbb{R}^{|\mathcal{I}_l|} $$
where $|\mathcal{I}_l|$ is the number of classes in the layer. 

However, the label-attention model is unaware of the overall hierarchical structure. Therefore, we ensemble it with another method. We refer to HiTIN \cite{Zhu2023HiTINHT} for the implementation of a state-of-the-art HTC global approach. In this method, a tree network is constructed based on the simplified original taxonomy structure, and the messages are propagated bottom-up in an isomorphism manner, which complements the label-attention model used. The embedding for leaf nodes are obtained by broadcasting the text representation $H$. After the tree isomorphism network propagation, all embedding from all layers are aggregated to form single embedding, and a classification layer is used to obtain the logits $H_{global}$ of all tree nodes. The logits are then split by the number of classes in each layer to obtain $H_{global}^l$.

The final class probabilities for each layer $P_l$ is then obtained by:
\vspace{-.3cm}
\[ P_l = softmax(H_{local}^l + H_{global}^l) \]
\vspace{-.5cm}

\begin{table}[t]
\centering
\scalebox{1}{
\begin{tabular}{||c c c c c||}
 \hline
 Market & Lang. & Intents & Train(ST) & Test(MT) \\ 
 \hline\hline
 BR & pt & 316 & 66k & 372 \\ 
 ID & id & 481 & 161k & 1145 \\
 MY & en,ms & 473 & 74k & 1417 \\
 PH & en,fil & 237 & 33k & 189 \\
 SG & en & 360 & 76k & 737 \\
 TH & th & 359 & 60k & 502 \\
 TW & zh-tw & 373 & 31k & 353 \\
 VN & vi & 389 & 178k & 525 \\ [1ex]
 \hline
\end{tabular}}
\caption{Multilingual dataset statistics for Single Turn (ST) and Multi-Turn (MT).}
\label{table:data_train_test}
\vspace{-.8cm}
\end{table}

\begin{table}[t]
\centering
\scalebox{1}{
\begin{tabular}{||c c c c c||} 
 \hline
 MKT & Model & $r_c$ & CL-Label & Accuracy \\ 
 \hline\hline
 SG & Naive Concat. & - & - & 60.52\% \\
 SG & Selective Concat. & - & - & 56.99\% \\
 SG & Llama2-7B & \ding{56}& \ding{56}& 56.24\% \\ 
 SG & Llama2-7B & \ding{52} & \ding{56}& 61.33\% \\
 SG & Domain-Llama2-7B & \ding{52} & \ding{56} & \textbf{63.23\%} \\
 ID & Naive Concat. & - & - & 60.61\% \\
 ID & Selective Concat. & - & - & \textbf{63.23\%} \\
 ID & Llama2-7B & \ding{52} & \ding{56} & 49.96\% \\
 ID & SeaLLM-7B-chat & \ding{52} & \ding{56} & 52.49\% \\
 ID & SeaLLM-7B-chat & \ding{52} & \ding{52} & 55.02\% \\ [1ex] 
 % prefix lm in ID is worse than selective expand, but better with naive expand
 \hline
\end{tabular}
}
\caption{Performance of LLM with symbol tuning approaches.}
\label{table:results_sft_llm}
\vspace{-.8cm}
\end{table}

\section{Experiments}
\label{sec:exp}

\subsection{Dataset} 
The dataset used in our experiments is derived from the conversation history of a large e-commerce platform. It includes user queries in the local languages of eight markets: Brazil (BR), Indonesia (ID), Malaysia (MY), Philippines (PH), Singapore (SG), Thailand (TH), Taiwan (TW), and Vietnam (VN), as detailed in Table~\ref{table:data_train_test}. Labeled data were manually annotated by local customer service teams, with only samples achieving label consistency across three independent taggers being selected to ensure quality.

Single-turn training data collected over years of business operations form the basis for supervised fine-tuning and in-context learning. For multi-turn evaluation, real online sessions are annotated by local customer service teams, with only the last query $q_n$ labeled in each session $\mathcal{Q}$. For preprocessing, we remove noisy annotations, standardize intents, and augment multi-turn sessions using dialogue state transition probabilities derived from chat logs.

\noindent\textbf{Symbol Tuning}. 
We perform symbol tuning on LLM for the SG and ID datasets, where SG mainly uses English while ID uses Bahasa Indonesia. The training data comprises a mix of existing single-turn samples and about 60k semi-automatically crafted multi-turn samples added to each market. Some are obtained by cleaning data on online chat logs to identify more accurate intents using an LLM with a few-shots of the chain-of-thought prompt. The rest are constructed by combining several dialogues sampled from the existing single-turn training dataset to form one session. The transition of intents in a session is calculated from the online chatlog.

\noindent\textbf{HTC with C-LARA}. 
70k of online chat logs are sampled for pseudo-labelling. After self-consistency checking, around 12\% of the data yield inconsistent results and are discarded from training. 1.5k samples are split from the pseudo-labeled data to serve as the validation set for early stopping. 

\subsection{Metrics}
The primary evaluation metric is the accuracy of predicted labels for the final query $q_n$ in each conversation session $\mathcal{Q}$. Metrics accounting for class imbalance were not considered, as the sampled sessions reflect the distribution of online traffic across intents, providing a realistic approximation of live performance.

\subsection{Implementation Details}
\subsubsection*{Symbol Tuning on LLM} \label{experiment:sft_details} 
FastChat framework is used to fine-tune 7B LLMs using LoRA method on their $q\_proj$, $v\_proj$, $o\_proj$, and $k\_proj$ modules with a learning rate of 2e-5 over 10 epochs. The 7B models used are Llama-2-7B (for SG) and SeaLLM-7B-chat (for ID) on Hugging Face. Before the models are fine-tuned on the multi-turn intent recognition task, they are further pre-trained on ShareGPT dataset with the same setting above, and the weights are then merged. For the sake of simplicity, we will refer to the LLMs further pre-trained on ShareGPT dataset as base models. During training for intent classification task, loss is calculated on all the model output including those after history queries. During inference, greedy decoding strategy is used to generate the target $r$ part, the prefix "The intent title is " is not generated but instead appended at the end of the prompt. When the generated label has no exact match with any $r$ in $\mathcal{I}$, gestalt string matching is used to find the closest one.

\subsubsection*{HTC with C-LARA} 
The in-house Hierarchical Text Classification (HTC) model is a BERT-based model fine-tuned using the combination of the pseudo-labeled multi-turn data and existing single-turn data, as shown in Section 4.1. We use AdamW to finetune the HTC with a learning rate of 5e-6. All tests are run on a single Nvidia V100 GPU card with 32GB of GPU memory.

% TODO Remove "Pipeline" column? "Model" -> "Method"
\begin{table*}[t]
\centering
\scalebox{.85}{
\begin{tabular}{||l l l c | c c c c c c c c c ||} 
 \hline
 Pipeline & Model & Prompt & Self-Consistency & BR & ID & MY & PH & SG & TH & TW & VN & avg \\ [0.5ex] 
 \hline\hline
 Fine-tuning & Single-turn & - & - & 30.98\% & 52.14\% & 56.81\% & 40.21\% & 51.13\% & 52.99\% & 58.07\% & 65.90\% & 53.76\% \\
 Fine-tuning & Naive Concat. & - & - & 50.81\% & 60.61\% & 57.02\% & 47.62\% & 60.52\% & 56.97\% & 65.44\% & 76.95\% & 60.08\% \\
 Fine-tuning & Selective Concat. & - & - & 52.69\% & 63.23\% & 60.20\% & 51.32\% & 56.99\% & 57.77\% & 64.02\% & 74.10\% & 60.97\% \\ 
 LARA & Vicuna-13B & $\mathcal{P}$ & \ding{56}& 52.69\% & 61.48\% & 65.42\% & 54.50\% & 65.26\% & 60.96\% & 67.14\% & 77.90\% & 64.18\% \\
 C-LARA & Vicuna-13B & $\mathcal{P}$ & \ding{52} & 55.38\% & 63.58\% & 65.00\% & 54.50\% & 66.21\% & 63.75\% & 71.10\% & 79.24\% & 65.52\% \\
 LARA & Vicuna-13B & $\mathcal{P}_{symbolic}$ & \ding{56}& 51.88\% & 60.00\% & 64.57\% & 53.97\% & 65.26\% & 58.96\% & 65.44\% & 74.67\% & 62.92\% \\
 C-LARA & Vicuna-13B & $\mathcal{P}_{symbolic}$ & \ding{52} & 54.57\% & 62.62\% & 65.56\% & 50.79\% & 66.76\% & 62.95\% & 69.97\% & 76.76\% & 64.94\% \\
 LARA & Vicuna-13B & $\mathcal{P}_{prepend}$ & \ding{56}& 54.03\% & 61.75\% & 64.50\% & 53.44\% & 65.94\% & 61.55\% & 66.86\% & 75.81\% & 63.97\% \\
 C-LARA & Vicuna-13B & $\mathcal{P}_{prepend}$ & \ding{52} & 53.76\% & 63.84\% & 65.70\% & 52.91\% & \textbf{68.11\%} & 63.15\% & 69.97\% & 78.48\% & 65.65\% \\
 LARA & Vicuna-13B & $\mathcal{P}_{formatted}$ & \ding{56}& 55.65\% & 62.88\% & 64.71\% & \textbf{55.03\%} & 65.40\% & 61.95\% & 66.86\% & 78.10\% & 64.64\% \\
 LARA-PL & Vicuna-13B & $\mathcal{P}_{formatted}$ & \ding{56}& \textbf{55.91\%} & 64.19\% & 64.43\% & 49.21\% & 66.49\% & 61.95\% & 69.41\% & \textbf{81.14\%} & 65.29\% \\
 C-LARA & Vicuna-13B & $\mathcal{P}_{formatted}$ & \ding{52} & \textbf{55.91\%} & \textbf{65.33\%} & \textbf{66.27\%} & 51.85\% & 67.16\% & \textbf{63.35\%} & \textbf{72.80\%} & 78.86\% & \textbf{66.35\%} \\[1ex]
 \hline
\end{tabular}
}
\caption{Performance of C-LARA compared to baselines, the average here is weighted on the number of test samples in each market. The results illustrate that C-LARA with formatted prompts achieves the best average accuracy (66.35\%) across all markets. The results validate our approach's effectiveness in both English and non-English markets, with significant improvements over baseline methods.}
\label{table:results_lara}
\vspace{-0.75cm}
\end{table*}

\subsection{Baseline settings}
For a fair comparison, we adopt three methods fine-tuned on HTC model ($\mathcal{M}_c$) as our global baselines across two methods:

\begin{enumerate}
\item\textbf{Single-turn method}: where only the last query of a session is considered by $\mathcal{M}_c$;

\item\textbf{Naive concatenation}: all queries are concatenated together before being fed into $\mathcal{M}_c$; 

\item\textbf{Selective concatenation}: where a concatenation selection model is trained to select the most suitable historical query with the last query to serve as the input to $\mathcal{M}_c$.
\end{enumerate}

\subsubsection*{ST on LLM}
% \noindent\textbf{Different Base Models.} 
In SG, except Llama2-7B, we also tried to continue pre-training the base models on in-domain corpus to strengthen the language understanding of local languages and the corresponding slang used, as humans usually converse with the chatbot in a non-formal way. We term the domain specific base model as \textbf{Domain-Llama2-7B}. In ID, we switched Llama2-7B model to \textbf{SeaLLM-7B-chat} \cite{nguyen2024seallmslargelanguage} which was introduced specifically for languages in South East Asia.

The ST approach was adapted for supervised intent recognition using compressed generation targets ($r_c$) and cross-lingual labels (CL\_label). These adjustments optimized performance by simplifying the generative task while maintaining semantic integrity. Comparisons with baseline methods in Table~\ref{table:results_sft_llm} show that ST achieves competitive results in English markets but faces challenges in non-English settings due to limitations in pre-training for low-resource languages.

\subsubsection*{HTC with C-LARA}
This experiment uses Vicuna-13B as our base model for pseudo-labeling within LARA and C-LARA. We designed three pipelines with four prompt templates in \cite{junhua2024lara} to demonstrate that using C-LARA for pseudo-labeling can effectively improve the HTC model's performance in multi-turn classification tasks. The detailed introduction is listed as follows:
\begin{itemize}
    \item \textbf{LARA}: Using LARA directly as a classifier.
    \item \textbf{LARA-PL}: Using LARA as a naive pseudo-labeling tool and fine-turn HTC model with generated data.
    \item \textbf{C-LARA}: Useing C-LARA to filter out the noise and generate high-quality data to fine-tune the HTC model.
\end{itemize}

\subsection{Offline Experiments}

\subsubsection*{Symbol Tuning on LLM}

Table~\ref{table:results_sft_llm} illustrates the effectiveness of Symbol Tuning (ST) on LLMs. Compressing the generation target $r$ reduces task complexity and improves accuracy by 5.09\% in the SG market. This compression also mitigates hallucination, reducing instances of unmatched generated labels from 2.5\% to 0\%.

Interestingly, this technique also stopped LLM hallucination, i.e. generating label with no match in the $\mathcal{I}$. The hallucination rate without using compressed $r$ is about 2.5\%. In ID, which is a non-English market, we find that \textbf{cross-lingual label} which changes the generation target to English rather than in the local language also improved the performance by 2.53\%. Using \textbf{different base models} which were trained specifically on the in-domain corpus or for the local language also proves to be useful. Domain-Llama2-7B improves the performance by 1.90\% in SG while SeaLLM-7B-chat improves the performance by 2.53\% in ID compared to Llama2-7B. While the ST approach outperforms the baselines in English market, it still leaves a lot to be desired in non-English market. This phenomenon may arise as a result of the ST approach employed for non-dominant languages during pre-training, which necessitates a greater quantity or higher quality of data to achieve satisfactory performance in a task that was not included in the pre-training phase. This is particularly true when the model lacks knowledge pertaining to domain intents. 

\subsubsection*{Pseudo-labeling using C-LARA}

As demonstrated in Table~\ref{table:results_lara}, C-LARA improves pseudo-label quality through self-consistency validation, resulting in a 1.06\% performance gain over LARA. This validation process identifies and removes approximately 12\% of inconsistent samples, ensuring high-quality synthetic labels. While this approach requires additional offline training resources, it significantly lowers deployment costs by relying on a single, lightweight classification model.

This most probably can be attributed to the advantages of the discriminative method in classification tasks, as training process also exposed the model to the comprehensive high quality single-turn dataset. Besides, the pre-trained model used for $\mathcal{M}_c$ was also pre-trained specifically on the in-domain multi-lingual corpus, making it a strong suit for our multilingual e-commerce setting. C-LARA’s integration of self-consistency within the pseudo-labeling pipeline significantly enhances the quality of synthetic labels, resulting in a 1.06\% improvement in performance, as indicated in the last row of Table \ref{table:results_lara}. When the LLM lacks confidence in its ICL responses, minor changes in the input prompt can significantly alter the output. This method effectively identifies potential inaccuracies in ICL outputs for black-box models where direct output scores are unavailable. Most importantly, this approach, while requiring longer offline training time, significantly reduces deployment costs to just one small classification model.

\subsection{Online Deployment Evaluation}
\subsubsection*{ST on LLM.} Using the LMDeploy framework, LARA weights were merged with the 7B base model, enabling faster inference times. Deployed on a single 32GB V100 GPU, the Symbol Tuning (ST) approach achieved an average latency of 170ms at 0.5 QPS in the SG market. In contrast, C-LARA models converted to ONNX format (1.1GB per model) achieved an average latency of 80ms at 1 QPS on an 8-core CPU machine with 16GB memory, demonstrating superior scalability and cost-efficiency.

\subsubsection*{C-LARA} 
\begin{figure}[t]
    \centering
    \includegraphics[width=\linewidth]{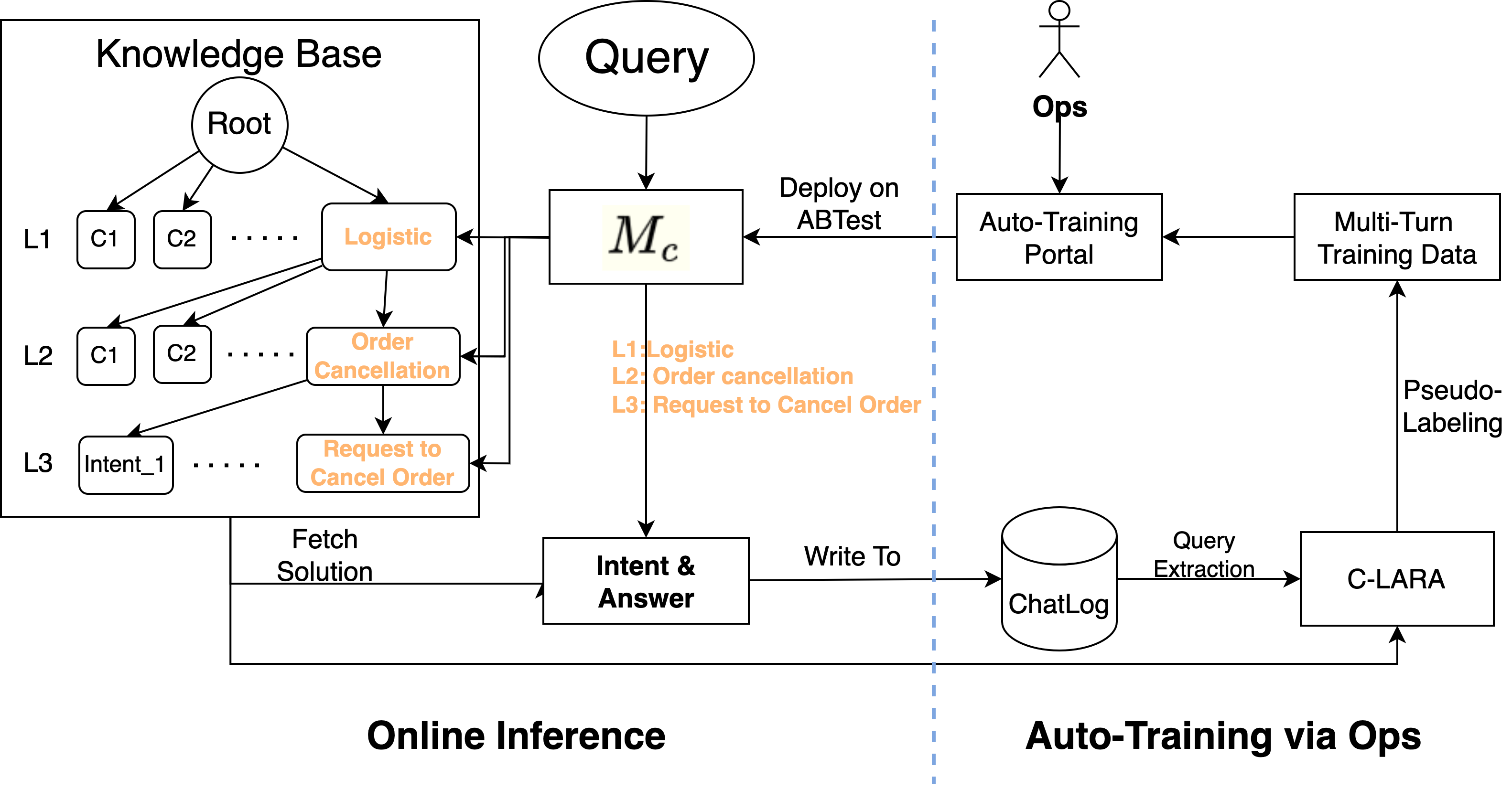}
    \vspace{-.7cm}
    \caption{Online Deployment of Multi-turn Intent Classification model demonstrates our production architecture integrating C-LARA for automated training data generation. The system handles real-time inference while continuously improving through automated training.}
    \label{fig:clara}
    \vspace{-.5cm}
\end{figure}

We deploy C-LARA across all eight markets. The models were first converted to ONNX format, reducing their size to 1.1GB. Deployed on an 8-core CPU machine with 16GB memory, C-LARA achieved an average latency of 80ms at 1 QPS, which is less than half the latency of the ST on LLM method. This deployment significantly reduced both costs and complexity, making it more scalable for industrial applications. Due to its versatility, an Auto-Training Portal (ATP) ecosystem is built around the LARA-PL method (Fig. \ref{fig:clara}). ATP enables seamless and continuous improvements for the chatbot’s multi-turn intent recognition system. Using online chat logs, local operations teams can update the Knowledge Base (KB) by adding new intents and crafting example queries. Subsequently, they can trigger C-LARA for pseudo-labeling multi-turn chat logs, generating data to train lightweight models. Once training is complete, the models are deployed through the portal for online A/B testing, creating an iterative cycle of improvement. For fair comparisons, the version of the KB (intents and single-turn training data) was kept consistent across control and test groups.

% Plese refer to this paper: https://arxiv.org/pdf/2408.01928
\subsection{Online Performance}
We leverage the following two metrics:

\begin{enumerate}
  \item \textbf{Resolution rate (RR)} which is measured by the rate of user completing the answer flow, not transferring to live agent, and not giving bad rating to the answer. 
  \item \textbf{Customer Service Satisfaction (SCSAT)} where users will be asked about their satisfaction towards our chatbot for chatbot only sessions (no intervention from live agents). The score is calculated by \textit{\# good rated sessions}/(\textit{\# good rated sessions} + \textit{\# bad rated sessions}).
\end{enumerate}

We use the \textbf{selective concatenation} method as the baseline for all experiments, with paired t-test to evaluate statistical significance.

\subsubsection*{ST on LLM} 
In the SG market, ST on LLMs was deployed to 50\% of online traffic for three weeks, yielding approximately 14k chatbot sessions per group. The test group exhibited a +2.19\% improvement in Customer Service Satisfaction (SCSAT), but Resolution Rate (RR) declined by -0.11\%. Neither result was statistically significant, indicating limited benefits from ST given its resource-intensive nature.

\subsubsection*{C-LARA} 
For C-LARA, only the multi-turn dialogue model was replaced, while single-turn models remained unchanged. Aggregated results from over 108k chatbot sessions per group showed statistically significant improvements: Resolution Rate (RR) increased by +0.78\% and Customer Service Satisfaction (SCSAT) by +1.39\% (p-value < 0.05). These gains translate to overall session improvements of RR +0.47\% and SCSAT +0.84\%, as multi-turn dialogues comprise 60.60\% of total sessions.

Furthermore, adding pseudo-labeled multi-turn data enhanced single-turn intent recognition. Substituting single-turn dialogue models with C-LARA models yielded an RR improvement of +0.06\% and a statistically significant SCSAT increase of +0.27\%. 

\begin{table*}[t]
\centering
\scalebox{1}{
\begin{tabular}{||l c | c c c c c c c c c ||} 
 \hline
 Prompt & Self-Consistency & BR & ID & MY & PH & SG & TH & TW & VN & avg \\ [0.5ex] 
 \hline\hline
  $\mathcal{P}$ & \ding{56}& 52.69\% & 61.48\% & 65.42\% & 54.50\% & 65.26\% & 60.96\% & 67.14\% & 77.90\% & 64.18\% \\
  $\mathcal{P}$ &  \ding{52} & 58.59\% & 68.13\% & 69.93\% & 56.44\% & 69.58\% & 66.75\% & 71.30\% & 81.14\% & 69.11\% \\
  $\mathcal{P}_{symbolic}$ & \ding{56}& 51.88\% & 60.00\% & 64.57\% & 53.97\% & 65.26\% & 58.96\% & 65.44\% & 74.67\% & 62.92\% \\
  $\mathcal{P}_{symbolic}$ &  \ding{52} & 56.63\% & 64.71\% & 68.48\% & 55.19\% & 68.77\% & 65.59\% & 71.61\% & 78.02\% & 67.27\% \\
  $\mathcal{P}_{prepend}$ & \ding{56}& 54.03\% & 61.75\% & 64.50\% & 53.44\% & 65.94\% & 61.55\% & 66.86\% & 75.81\% & 63.97\% \\
  $\mathcal{P}_{prepend}$ &  \ding{52} & 59.49\% & 66.08\% & 68.60\% & 55.90\% & 68.85\% & 68.19\% & 71.79\% & 81.36\% & 68.43\% \\
  $\mathcal{P}_{formatted}$ & \ding{56}& 55.65\% & 62.88\% & 64.71\% & 55.03\% & 65.40\% & 61.95\% & 66.86\% & 78.10\% & 64.64\% \\
  $\mathcal{P}_{formatted}$ &  \ding{52} & 59.24\% & 67.01\% & 69.47\% & 56.79\% & 68.81\% & 68.69\% & 72.35\% & 82.93\% & 69.12\% \\[1ex]
 \hline
\end{tabular}
}
\caption{Precision of C-LARA variants after filtering inconsistent predictions demonstrates the effectiveness of self-consistency checking across different prompt types (P, $P_symbolic$, $P_prepend$, $P_formatted$). Across all prompt types, accuracy improves by approximately 4-5\%, with $ \mathcal{P}_{formatted} $ achieving the highest precision (69.12\%). These results validate the robustness of self-consistency as a filtering strategy.}
\label{table:ablation_self_consistency}
\vspace{-0.7cm}
\end{table*}

\begin{table}[t]
\centering
\scalebox{1}{
\begin{tabular}{||l l | c c c ||} 
 \hline
 Method & Prompt & ID & SG & avg \\ [0.5ex] 
 \hline\hline
 LLM ST & - & 58.17\% & 63.23\% & 60.15\% \\
 C-LARA & $\mathcal{P}$ & 60.44\% & 64.31\% & 61.96\% \\
 C-LARA & $\mathcal{P}_{formatted}$ & 59.83\% & 64.04\% & 61.48\% \\[1ex]
 \hline
\end{tabular}
}
\caption{Results of LARA using 7B LLM.}
\label{table:ablation_model_size}
\vspace{-0.7cm}
\end{table}

\section{Ablation Study}

\subsection{ Effect of Target Length}
We investigate how the amount of information in ST generation target affects the intent recognition performance using two rather extreme approaches and their conversation semantic fluidity. 

% TODO or rephrase as multi-task? The way of adding auxiliary task to improve performance, other possibilities include during further pre-training stage, during fine-tuning but separate tasks into multiple targets/diff training sample
\subsubsection { Longer Target Length } To achieve this, the model is trained to summarize all queries in $\mathcal{Q}$ before outputting the target $r$. For instance, the new output format of model will be ``\textbf{You are asking about \{$summary$\}. So,} the intent title is \{$r_n$\}". The rationale is to utilize the summarization ability of LLMs to better understand the context. For our training data, the summaries are obtained by prompting the original LLM backbones in a zero-shot manner. We chose this over increasing the length of $r$ statically to impose more information on the model's generation target. Table~\ref{table:ablation_longer_target} demonstrates the impact of increasing the target length in Symbol Tuning (ST). Extending the generation target to include query summaries decreases performance by 3.82\%. While this approach enhances semantic coherence, excessive information overloads the model, reducing its ability to focus on the core intent classification task.

\begin{table}[t]
\centering
\scalebox{1.0}{
\begin{tabular}{||l l c c||} 
\hline
 MKT & Model & Longer Target Length & Accuracy \\ 
 \hline\hline
 SG & Llama2-7B & \ding{56} & 54.82\% \\ 
 SG & Llama2-7B & \ding{52} & 51.02\% \\ [1ex] 
 % prefix lm in ID is worse than selective expand, but better with naive expand
 \hline
\end{tabular}
}
\caption{Effect of Longer Target Length on LLM ST Classification Performance: lengthening targets results in a performance drop by 3.82\%.}
\label{table:ablation_longer_target}
\vspace{-.7cm}
\end{table}

\begin{table}[t]
\centering
\scalebox{1.0}{
\begin{tabular}{||l l c c||} 
 \hline
 MKT & Model & Shorter Target Length & Accuracy \\ 
 \hline\hline
 ID & SeaLLM-7B & \ding{56} & 55.02\% \\ 
 ID & SeaLLM-7B & \ding{52} & 46.11\% \\ [1ex] 
 \hline
\end{tabular}
}
\caption{Effect of Shorter Target Length on LLM ST Classification Performance: reducing semantically rich targets into symbols costs a drastic performance drop by 8.91\%.}
\label{table:ablation_shorter_target}
\vspace{-.8cm}
\end{table}

% \begin{table}[t]
% \centering
% \scalebox{1}{
% \begin{tabular}{||l l c c c||} 
%  \hline
%  MKT & Model & Masking & Pre-training & Accuracy \\ 
%  \hline\hline
%  ID & SeaLLM-7B & \ding{56} & \ding{52} & 55.02\% \\ 
%  ID & SeaLLM-7B & \ding{52} & \ding{56} & 52.14\% \\
%  ID & SeaLLM-7B & \ding{52} & \ding{52} & 56.86\% \\ [1ex]
%  \hline
% \end{tabular}
% }
% \caption{Performance of Prefix Language Model Masking Approaches.}
% \label{table:ablation_prefix_modeling}
% \end{table}

\subsubsection { Shorter Target Length } The approach of compressing $r$ was inspired by \cite{wei2023symboltuningimprovesincontext}. Hence, we also tried to replace $r$s with completely meaningless symbols, while keeping the generation prefix of ``The intent title is ". Compressing target labels to purely symbolic representations results in a significant 8.91\% performance drop, as shown in Table~\ref{table:ablation_shorter_target}. This highlights the importance of preserving semantic richness in target labels for generative fine-tuning. Effective compression methods must retain key information from the original labels to avoid loss in classification accuracy.. Thus, when compressing $r$s, it is important to choose a method that can preserve the information in original $r$s as much as possible.

\subsection{ Impact of Self-consistency in MTIC }
Using our multi-turn test sets, we evaluate the performance of MTIC with and without self-consistency checking. We remove the samples with inconsistent outputs and calculate the precision of the remaining samples. On average, 12\% of test samples will be removed in each market. Incorporating self-consistency checking into MTIC evaluations improves accuracy across all prompt variations, as shown in Table~\ref{table:ablation_self_consistency}. By removing approximately 12\% of test samples with inconsistent outputs, this method effectively filters out erroneous predictions, ensuring higher-quality pseudo-labels and more reliable results. This ensures the quality of pseudo-labels.

\subsection{ Effect of Model Size }
For fair comparison between LLM ST and C-LARA, we use vicuna-7b-v1.5 as the base model with prompt $\mathcal{P}$ and $\mathcal{P}_{formatted}$, without self-consistency checking. The results of LLM ST method are taken from the best of each market reported in this paper, including base models pre-trained on in-domain corpus, so it should have the advantage over Vicuna-7B-v1.5. Table~\ref{table:ablation_model_size} compares C-LARA and LLM ST using models of the same size (Vicuna-7B-v1.5) without self-consistency checking. Despite the simpler pipeline, C-LARA consistently outperforms LLM ST, avoiding the complexity of multi-turn sample crafting. However, smaller models exhibit reduced instruction-following capabilities, as demonstrated by the lower performance of $ \mathcal{P}_{formatted} $ compared to $ \mathcal{P} $. One interesting observation here is that the performance of C-LARA when using $\mathcal{P}_{formatted}$ is now lower than $\mathcal{P}$. LLMs of smaller size could be weaker in instruction following, and in this sense the semantic meaning of the labels in demonstrations are more critical. Prepending meaningless characters before labels can negatively affect the understanding of labels for smaller LLMs.

% \noindent \textbf{Summary.} The ablation studies reveal key insights into the role of target length, self-consistency, and model size in optimizing MTIC performance. While shorter and semantically rich targets enhance accuracy, self-consistency further ensures high-quality pseudo-labels, validating the effectiveness of the proposed C-LARA framework.
\section{Related Work}
\subsection{Synthetic Data Generation} 
The scarcity of annotated dialogue data, particularly in low-resource languages, has driven research into synthetic data generation. \citet{borisov_et_al_2022} proposed a method leveraging auto-regressive generative models to create realistic tabular datasets, highlighting their utility in data augmentation. Similarly, \citet{li_et_al_2023} demonstrated that synthetic data generated by LLMs can significantly enhance model performance in classification tasks. Additionally, \citet{tang_et_al_2024} utilized synthetic data to craft challenging examples for fact-checking, improving the factual accuracy of LLM outputs.

\subsection{Modeling Multi-turn Dialogue Context}
Multi-turn dialogue modeling is essential for dialogue understanding tasks. Early methods used bidirectional contextual LSTMs~\cite{ghosal2021exploring} to capture context-aware utterance representations for tasks such as MultiWOZ intent classification~\cite{budzianowski2018multiwoz}. Other approaches, such as multi-channel graph convolutional networks, were applied to query classification in E-commerce~\cite{yuan2024semi}. 

Recent advancements leverage pre-trained language models (PLMs) as sentence encoders~\cite{shen2021directed}, particularly for emotion recognition in conversations (ERC). For instance, \citet{lee2022compm} encoded both context and speaker memory using PLMs, while \citet{qin2023bert} incorporated multi-turn information from utterances and dialogue structure through fine-tuning. Despite their effectiveness, these methods depend heavily on multi-turn training datasets, which are difficult to acquire in real-world e-commerce settings~\cite{liu2024responsible}. In contrast, our approach employs LLMs within an augmentation-based pipeline to generate multi-turn data, enabling zero-shot intent classification using smaller models.

\subsection{LLM on text classification}
Recent studies have explored the applicability of LLMs across various domains. \citet{chae2023large} investigated LLMs for sociological text classification, demonstrating their potential in social science research. In financial intent detection, \citet{loukas2023making} analyzed the trade-offs between performance and cost when using LLMs for text classification. \citet{liu2024towards} employed GPT-4o to perform zero-shot classification on multi-level semi-structured text with retrieval augmentation. \citet{wei2023empirical} highlighted the benefits of fine-tuning LLMs on domain-specific datasets, improving performance in legal document review. \citet{wei2023symboltuningimprovesincontext} introduced symbol tuning, where natural language labels were replaced with unrelated symbols during fine-tuning to enhance classification. Our work differs by compressing longer intent labels into semantically meaningful phrases, enabling easier generation and improving accuracy for tasks with a large number of classes.

\section{Conclusion}
Multi-turn intent classification plays a critical role in modern dialogue systems. Unlike typical classification tasks, real-world intent classification often involves varying intent lengths, posing unique challenges. In this work, we introduced Symbol Tuning to fine-tune large language models (LLMs) with compressed intents. Our experiments demonstrated that shortening intents improved accuracy by 5.09\% compared to using original intents.

Additionally, we proposed C-LARA, an augmentation-based pipeline for generating high-quality multi-turn datasets using self-consistency validation. Training smaller models with pseudo-labeled data generated by C-LARA yielded a 1.06\% average performance improvement. Empirically, C-LARA significantly reduces annotation costs by automating pseudo-labeling based on the user’s latest utterance in dialogue history, improving model iteration efficiency. Furthermore, training smaller models offers computational efficiency, enabling scalable deployment and online inference.

\noindent \textbf{Future Work.} Moving forward, we aim to incorporate features such as user profiles and order history into C-LARA to support more diverse dialogue tasks. We also plan to explore cross-lingual transfer and advanced tokenization techniques to enhance performance in low-resource languages.

% \begin{acks}
% Some acknowledgement
% \end{acks}

\bibliographystyle{ACM-Reference-Format}
\bibliography{custom}


\begin{thebibliography}{26}

%%% the Web address.

\ifx \showCODEN    \undefined \def \showCODEN     #1{\unskip}     \fi
\ifx \showDOI      \undefined \def \showDOI       #1{#1}\fi
\ifx \showISBNx    \undefined \def \showISBNx     #1{\unskip}     \fi
\ifx \showISBNxiii \undefined \def \showISBNxiii  #1{\unskip}     \fi
\ifx \showISSN     \undefined \def \showISSN      #1{\unskip}     \fi
\ifx \showLCCN     \undefined \def \showLCCN      #1{\unskip}     \fi
\ifx \shownote     \undefined \def \shownote      #1{#1}          \fi
\ifx \showarticletitle \undefined \def \showarticletitle #1{#1}   \fi
\ifx \showURL      \undefined \def \showURL       {\relax}        \fi
% invisible to TeX
\providecommand\bibfield[2]{#2}
\providecommand\bibinfo[2]{#2}
\providecommand\natexlab[1]{#1}
\providecommand\showeprint[2][]{arXiv:#2}

\bibitem[Borisov et~al\mbox{.}(2022)]%
        {borisov_et_al_2022}
\bibfield{author}{\bibinfo{person}{Vadim Borisov}, \bibinfo{person}{Kathrin Seßler}, \bibinfo{person}{Tobias Leemann}, \bibinfo{person}{Martin Pawelczyk}, {and} \bibinfo{person}{Gjergji Kasneci}.} \bibinfo{year}{2022}\natexlab{}.
\newblock \showarticletitle{Language Models Are Realistic Tabular Data Generators}.
\newblock \bibinfo{journal}{\emph{ArXiv}}  \bibinfo{volume}{abs/2210.06280} (\bibinfo{year}{2022}).
\newblock


\bibitem[Budzianowski et~al\mbox{.}(2018)]%
        {budzianowski2018multiwoz}
\bibfield{author}{\bibinfo{person}{Pawe{\l} Budzianowski}, \bibinfo{person}{Tsung-Hsien Wen}, \bibinfo{person}{Bo-Hsiang Tseng}, \bibinfo{person}{Inigo Casanueva}, \bibinfo{person}{Stefan Ultes}, \bibinfo{person}{Osman Ramadan}, {and} \bibinfo{person}{Milica Ga{\v{s}}i{\'c}}.} \bibinfo{year}{2018}\natexlab{}.
\newblock \showarticletitle{MultiWOZ--a large-scale multi-domain wizard-of-oz dataset for task-oriented dialogue modelling}.
\newblock \bibinfo{journal}{\emph{arXiv preprint arXiv:1810.00278}} (\bibinfo{year}{2018}).
\newblock


\bibitem[Chae and Davidson(2023)]%
        {chae2023large}
\bibfield{author}{\bibinfo{person}{Youngjin Chae} {and} \bibinfo{person}{Thomas Davidson}.} \bibinfo{year}{2023}\natexlab{}.
\newblock \showarticletitle{Large language models for text classification: From zero-shot learning to fine-tuning}.
\newblock \bibinfo{journal}{\emph{Open Science Foundation}} (\bibinfo{year}{2023}).
\newblock


\bibitem[Ghosal et~al\mbox{.}(2021)]%
        {ghosal2021exploring}
\bibfield{author}{\bibinfo{person}{Deepanway Ghosal}, \bibinfo{person}{Navonil Majumder}, \bibinfo{person}{Rada Mihalcea}, {and} \bibinfo{person}{Soujanya Poria}.} \bibinfo{year}{2021}\natexlab{}.
\newblock \showarticletitle{Exploring the role of context in utterance-level emotion, act and intent classification in conversations: An empirical study}. In \bibinfo{booktitle}{\emph{Findings of the Association for Computational Linguistics: ACL-IJCNLP 2021}}. \bibinfo{pages}{1435--1449}.
\newblock


\bibitem[Lee and Lee(2022)]%
        {lee2022compm}
\bibfield{author}{\bibinfo{person}{Joosung Lee} {and} \bibinfo{person}{Wooin Lee}.} \bibinfo{year}{2022}\natexlab{}.
\newblock \showarticletitle{CoMPM: Context Modeling with Speaker’s Pre-trained Memory Tracking for Emotion Recognition in Conversation}. In \bibinfo{booktitle}{\emph{Proceedings of the 2022 Conference of the North American Chapter of the Association for Computational Linguistics: Human Language Technologies}}. \bibinfo{pages}{5669--5679}.
\newblock


\bibitem[Li et~al\mbox{.}(2023)]%
        {li_et_al_2023}
\bibfield{author}{\bibinfo{person}{Zhuoyan Li}, \bibinfo{person}{Hangxiao Zhu}, \bibinfo{person}{Zhuoran Lu}, {and} \bibinfo{person}{Ming Yin}.} \bibinfo{year}{2023}\natexlab{}.
\newblock \showarticletitle{Synthetic Data Generation with Large Language Models for Text Classification: Potential and Limitations}.
\newblock \bibinfo{journal}{\emph{Proceedings of the 2023 Conference on Empirical Methods in Natural Language Processing (EMNLP)}} (\bibinfo{year}{2023}).
\newblock


\bibitem[Liu and Fu(2024)]%
        {liu2024responsible}
\bibfield{author}{\bibinfo{person}{Junhua Liu} {and} \bibinfo{person}{Bin Fu}.} \bibinfo{year}{2024}\natexlab{}.
\newblock \showarticletitle{Responsible Multilingual Large Language Models: A Survey of Development, Applications, and Societal Impact}.
\newblock \bibinfo{journal}{\emph{ArXiv}} (\bibinfo{year}{2024}).
\newblock


\bibitem[Liu et~al\mbox{.}(2024a)]%
        {liu2024towards}
\bibfield{author}{\bibinfo{person}{Junhua Liu}, \bibinfo{person}{Kwan~Hui Lim}, {and} \bibinfo{person}{Roy Ka-Wei Lee}.} \bibinfo{year}{2024}\natexlab{a}.
\newblock \showarticletitle{Towards Objective and Unbiased Decision Assessments with LLM-Enhanced Hierarchical Attention Networks}.
\newblock \bibinfo{journal}{\emph{arXiv preprint arXiv:2411.08504}} (\bibinfo{year}{2024}).
\newblock


\bibitem[Liu et~al\mbox{.}(2024b)]%
        {junhua2024lara}
\bibfield{author}{\bibinfo{person}{Junhua Liu}, \bibinfo{person}{Yong~Keat Tan}, \bibinfo{person}{Bin Fu}, {and} \bibinfo{person}{Kwan~Hui Lim}.} \bibinfo{year}{2024}\natexlab{b}.
\newblock \showarticletitle{LARA: Linguistic-Adaptive Retrieval-Augmentation for Multi-Turn Intent Classification}.
\newblock \bibinfo{journal}{\emph{Proceedings of the Empirical Methods in Natural Language Processing}} (\bibinfo{year}{2024}).
\newblock


\bibitem[Loukas et~al\mbox{.}(2023)]%
        {loukas2023making}
\bibfield{author}{\bibinfo{person}{Lefteris Loukas}, \bibinfo{person}{Ilias Stogiannidis}, \bibinfo{person}{Odysseas Diamantopoulos}, \bibinfo{person}{Prodromos Malakasiotis}, {and} \bibinfo{person}{Stavros Vassos}.} \bibinfo{year}{2023}\natexlab{}.
\newblock \showarticletitle{Making llms worth every penny: Resource-limited text classification in banking}. In \bibinfo{booktitle}{\emph{Proceedings of the Fourth ACM International Conference on AI in Finance}}. \bibinfo{pages}{392--400}.
\newblock


\bibitem[Nguyen et~al\mbox{.}(2024)]%
        {nguyen2024seallmslargelanguage}
\bibfield{author}{\bibinfo{person}{Xuan-Phi Nguyen}, \bibinfo{person}{Wenxuan Zhang}, \bibinfo{person}{Xin Li}, \bibinfo{person}{Mahani Aljunied}, \bibinfo{person}{Zhiqiang Hu}, \bibinfo{person}{Chenhui Shen}, \bibinfo{person}{Yew~Ken Chia}, \bibinfo{person}{Xingxuan Li}, \bibinfo{person}{Jianyu Wang}, \bibinfo{person}{Qingyu Tan}, \bibinfo{person}{Liying Cheng}, \bibinfo{person}{Guanzheng Chen}, \bibinfo{person}{Yue Deng}, \bibinfo{person}{Sen Yang}, \bibinfo{person}{Chaoqun Liu}, \bibinfo{person}{Hang Zhang}, {and} \bibinfo{person}{Lidong Bing}.} \bibinfo{year}{2024}\natexlab{}.
\newblock \bibinfo{title}{SeaLLMs -- Large Language Models for Southeast Asia}.
\newblock
\newblock
\showeprint[arxiv]{2312.00738}~[cs.CL]
\urldef\tempurl%
\url{https://arxiv.org/abs/2312.00738}
\showURL{%
\tempurl}


\bibitem[P{\v{r}}ib{\'a}{\v{n}} et~al\mbox{.}(2024)]%
        {pvribavn2024comparative}
\bibfield{author}{\bibinfo{person}{Pavel P{\v{r}}ib{\'a}{\v{n}}}, \bibinfo{person}{Jakub {\v{S}}m{\'\i}d}, \bibinfo{person}{Josef Steinberger}, {and} \bibinfo{person}{Adam Mi{\v{s}}tera}.} \bibinfo{year}{2024}\natexlab{}.
\newblock \showarticletitle{A comparative study of cross-lingual sentiment analysis}.
\newblock \bibinfo{journal}{\emph{Expert Systems with Applications}}  \bibinfo{volume}{247} (\bibinfo{year}{2024}), \bibinfo{pages}{123247}.
\newblock


\bibitem[Qin et~al\mbox{.}(2020)]%
        {qin2020dcr}
\bibfield{author}{\bibinfo{person}{Libo Qin}, \bibinfo{person}{Wanxiang Che}, \bibinfo{person}{Yangming Li}, \bibinfo{person}{Mingheng Ni}, {and} \bibinfo{person}{Ting Liu}.} \bibinfo{year}{2020}\natexlab{}.
\newblock \showarticletitle{Dcr-net: A deep co-interactive relation network for joint dialog act recognition and sentiment classification}. In \bibinfo{booktitle}{\emph{Proceedings of the AAAI conference on artificial intelligence}}, Vol.~\bibinfo{volume}{34}. \bibinfo{pages}{8665--8672}.
\newblock


\bibitem[Qin et~al\mbox{.}(2023)]%
        {qin2023bert}
\bibfield{author}{\bibinfo{person}{Xiangyu Qin}, \bibinfo{person}{Zhiyu Wu}, \bibinfo{person}{Tingting Zhang}, \bibinfo{person}{Yanran Li}, \bibinfo{person}{Jian Luan}, \bibinfo{person}{Bin Wang}, \bibinfo{person}{Li Wang}, {and} \bibinfo{person}{Jinshi Cui}.} \bibinfo{year}{2023}\natexlab{}.
\newblock \showarticletitle{Bert-erc: Fine-tuning bert is enough for emotion recognition in conversation}. In \bibinfo{booktitle}{\emph{Proceedings of the AAAI Conference on Artificial Intelligence}}, Vol.~\bibinfo{volume}{37}. \bibinfo{pages}{13492--13500}.
\newblock


\bibitem[Qu et~al\mbox{.}(2019)]%
        {Qu_2019}
\bibfield{author}{\bibinfo{person}{Chen Qu}, \bibinfo{person}{Liu Yang}, \bibinfo{person}{W.~Bruce Croft}, \bibinfo{person}{Yongfeng Zhang}, \bibinfo{person}{Johanne~R. Trippas}, {and} \bibinfo{person}{Minghui Qiu}.} \bibinfo{year}{2019}\natexlab{}.
\newblock \showarticletitle{User Intent Prediction in Information-seeking Conversations}. In \bibinfo{booktitle}{\emph{Proceedings of the 2019 Conference on Human Information Interaction and Retrieval}} \emph{(\bibinfo{series}{CHIIR ’19})}. \bibinfo{publisher}{ACM}.
\newblock
\urldef\tempurl%
\url{https://doi.org/10.1145/3295750.3298924}
\showDOI{\tempurl}


\bibitem[Rust et~al\mbox{.}(2021)]%
        {rust2021good}
\bibfield{author}{\bibinfo{person}{Phillip Rust}, \bibinfo{person}{Jonas Pfeiffer}, \bibinfo{person}{Ivan Vuli{\'c}}, \bibinfo{person}{Sebastian Ruder}, {and} \bibinfo{person}{Iryna Gurevych}.} \bibinfo{year}{2021}\natexlab{}.
\newblock \showarticletitle{How Good is Your Tokenizer? On the Monolingual Performance of Multilingual Language Models}. In \bibinfo{booktitle}{\emph{Proceedings of the 59th Annual Meeting of the Association for Computational Linguistics and the 11th International Joint Conference on Natural Language Processing (Volume 1: Long Papers)}}. \bibinfo{pages}{3118--3135}.
\newblock


\bibitem[Shen et~al\mbox{.}(2021)]%
        {shen2021directed}
\bibfield{author}{\bibinfo{person}{Weizhou Shen}, \bibinfo{person}{Siyue Wu}, \bibinfo{person}{Yunyi Yang}, {and} \bibinfo{person}{Xiaojun Quan}.} \bibinfo{year}{2021}\natexlab{}.
\newblock \showarticletitle{Directed Acyclic Graph Network for Conversational Emotion Recognition}. In \bibinfo{booktitle}{\emph{Proceedings of the 59th Annual Meeting of the Association for Computational Linguistics and the 11th International Joint Conference on Natural Language Processing (Volume 1: Long Papers)}}. \bibinfo{pages}{1551--1560}.
\newblock


\bibitem[Tang et~al\mbox{.}(2024)]%
        {tang_et_al_2024}
\bibfield{author}{\bibinfo{person}{Liyan Tang}, \bibinfo{person}{Philippe Laban}, {and} \bibinfo{person}{Greg Durrett}.} \bibinfo{year}{2024}\natexlab{}.
\newblock \showarticletitle{MiniCheck: Efficient Fact-Checking of LLMs on Grounding Documents}.
\newblock \bibinfo{journal}{\emph{ArXiv}}  \bibinfo{volume}{abs/2404.10774} (\bibinfo{year}{2024}).
\newblock


\bibitem[Wang et~al\mbox{.}(2024)]%
        {wang2024smart}
\bibfield{author}{\bibinfo{person}{Zhiqiang Wang}, \bibinfo{person}{Yiran Pang}, {and} \bibinfo{person}{Yanbin Lin}.} \bibinfo{year}{2024}\natexlab{}.
\newblock \showarticletitle{Smart Expert System: Large Language Models as Text Classifiers}.
\newblock \bibinfo{journal}{\emph{arXiv preprint arXiv:2405.10523}} (\bibinfo{year}{2024}).
\newblock


\bibitem[Wei et~al\mbox{.}(2023b)]%
        {wei2023empirical}
\bibfield{author}{\bibinfo{person}{Fusheng Wei}, \bibinfo{person}{Robert Keeling}, \bibinfo{person}{Nathaniel Huber-Fliflet}, \bibinfo{person}{Jianping Zhang}, \bibinfo{person}{Adam Dabrowski}, \bibinfo{person}{Jingchao Yang}, \bibinfo{person}{Qiang Mao}, {and} \bibinfo{person}{Han Qin}.} \bibinfo{year}{2023}\natexlab{b}.
\newblock \showarticletitle{Empirical study of LLM fine-tuning for text classification in legal document review}. In \bibinfo{booktitle}{\emph{2023 IEEE International Conference on Big Data (BigData)}}. IEEE, \bibinfo{pages}{2786--2792}.
\newblock


\bibitem[Wei et~al\mbox{.}(2023a)]%
        {wei2023symboltuningimprovesincontext}
\bibfield{author}{\bibinfo{person}{Jerry Wei}, \bibinfo{person}{Le Hou}, \bibinfo{person}{Andrew Lampinen}, \bibinfo{person}{Xiangning Chen}, \bibinfo{person}{Da Huang}, \bibinfo{person}{Yi Tay}, \bibinfo{person}{Xinyun Chen}, \bibinfo{person}{Yifeng Lu}, \bibinfo{person}{Denny Zhou}, \bibinfo{person}{Tengyu Ma}, {and} \bibinfo{person}{Quoc~V. Le}.} \bibinfo{year}{2023}\natexlab{a}.
\newblock \bibinfo{title}{Symbol tuning improves in-context learning in language models}.
\newblock
\newblock
\showeprint[arxiv]{2305.08298}~[cs.CL]
\urldef\tempurl%
\url{https://arxiv.org/abs/2305.08298}
\showURL{%
\tempurl}


\bibitem[Weld et~al\mbox{.}(2021)]%
        {weld2021survey}
\bibfield{author}{\bibinfo{person}{H. Weld}, \bibinfo{person}{X. Huang}, \bibinfo{person}{S. Long}, \bibinfo{person}{J. Poon}, {and} \bibinfo{person}{S.~C. Han}.} \bibinfo{year}{2021}\natexlab{}.
\newblock \bibinfo{title}{A survey of joint intent detection and slot-filling models in natural language understanding}.
\newblock
\newblock
\showeprint[arxiv]{2101.08091}~[cs.CL]


\bibitem[Wu et~al\mbox{.}(2021)]%
        {wu2021contextaware}
\bibfield{author}{\bibinfo{person}{Ting-Wei Wu}, \bibinfo{person}{Ruolin Su}, {and} \bibinfo{person}{Biing-Hwang Juang}.} \bibinfo{year}{2021}\natexlab{}.
\newblock \bibinfo{title}{A Context-Aware Hierarchical BERT Fusion Network for Multi-turn Dialog Act Detection}.
\newblock
\newblock
\showeprint[arxiv]{2109.01267}~[cs.CL]


\bibitem[Xu and Sarikaya(2014)]%
        {6853573}
\bibfield{author}{\bibinfo{person}{Puyang Xu} {and} \bibinfo{person}{Ruhi Sarikaya}.} \bibinfo{year}{2014}\natexlab{}.
\newblock \showarticletitle{Contextual domain classification in spoken language understanding systems using recurrent neural network}. In \bibinfo{booktitle}{\emph{2014 IEEE International Conference on Acoustics, Speech and Signal Processing (ICASSP)}}. \bibinfo{pages}{136--140}.
\newblock
\urldef\tempurl%
\url{https://doi.org/10.1109/ICASSP.2014.6853573}
\showDOI{\tempurl}


\bibitem[Yuan et~al\mbox{.}(2024)]%
        {yuan2024semi}
\bibfield{author}{\bibinfo{person}{Chunyuan Yuan}, \bibinfo{person}{Ming Pang}, \bibinfo{person}{Zheng Fang}, \bibinfo{person}{Xue Jiang}, \bibinfo{person}{Changping Peng}, {and} \bibinfo{person}{Zhangang Lin}.} \bibinfo{year}{2024}\natexlab{}.
\newblock \showarticletitle{A Semi-supervised Multi-channel Graph Convolutional Network for Query Classification in E-commerce}. In \bibinfo{booktitle}{\emph{Companion Proceedings of the ACM on Web Conference 2024}}. \bibinfo{pages}{56--64}.
\newblock


\bibitem[Zhu et~al\mbox{.}(2023)]%
        {Zhu2023HiTINHT}
\bibfield{author}{\bibinfo{person}{He Zhu}, \bibinfo{person}{Chong Zhang}, \bibinfo{person}{Junjie Huang}, \bibinfo{person}{Junran Wu}, {and} \bibinfo{person}{Ke Xu}.} \bibinfo{year}{2023}\natexlab{}.
\newblock \showarticletitle{HiTIN: Hierarchy-aware Tree Isomorphism Network for Hierarchical Text Classification}. In \bibinfo{booktitle}{\emph{Annual Meeting of the Association for Computational Linguistics}}.
\newblock
\urldef\tempurl%
\url{https://api.semanticscholar.org/CorpusID:258865236}
\showURL{%
\tempurl}


\end{thebibliography}

\appendix

\end{document}